\documentclass[11pt,a4paper]{article}
\usepackage[hyperref]{acl2019}
\usepackage{times}
\usepackage{latexsym}

\usepackage{url}
\usepackage{amsmath}
\usepackage{bbm}
\usepackage{graphicx}
\usepackage{booktabs}
\usepackage{tikz}
\usepackage{tikzscale}
\usepackage{tikz-dependency}

\aclfinalcopy
\def\aclpaperid{393}

\newcommand{\fig}[1]{Figure~\ref{fig:#1}}
\newcommand{\tab}[1]{Table~\ref{tab:#1}}
\newcommand{\eq}[1]{Eq.~\ref{eq:#1}}

\title{Adaptive Attention Span in Transformers}

\author{
Sainbayar Sukhbaatar \quad Edouard Grave \quad Piotr Bojanowski \quad Armand Joulin \\
  Facebook AI Research\\
  \texttt{\{sainbar,egrave,bojanowski,ajoulin\}@fb.com} \\
}

\date{}

\begin{document}
\maketitle

\begin{abstract}
We propose a novel self-attention mechanism that can learn its optimal attention span.
This allows us to extend significantly the maximum context size used in Transformer,
while maintaining control over their memory footprint and computational time.
We show the effectiveness of our approach on the task of character level language modeling, where we achieve state-of-the-art performances on \texttt{text8} and \texttt{enwiki8} by using a maximum context of $8$k characters.
\end{abstract}

\section{Introduction}

Language models are at the core of many NLP applications, like machine translation or dialogue.
Recently, much progress has been made by a new neural network called Transformer~\citep{vaswani2017attention}.
Part of its success is due to its ability to capture long term dependencies.
This is achieved by taking long sequences as inputs and explicitly compute the relations between every token via a mechanism called the ``self-attention'' layer~\citep{al2018character}.

While this layer allows for information to propagate across long distances, it has a computational and memory cost that scales quadratically with the size of the input sequence.
As a consequence, Transformers hardly scale to sequences of more than a thousand  tokens.
This is particularly problematic in the case of character level language modeling  where dependencies are often spread over a few thousands time steps.

In this work, we propose an alternative to the self-attention layer to reduce the computational burden of a Transformer.
Our layer learns its optimal context size, resulting in a network where each attention layer gathers information on their own context.
In practice, we observe that this leads to  Transformer with small context in the low-level layers and very large ones for the last layers.
With this modification, we are able to scale input sequences to more than $8$k tokens with no loss of performance, nor additional computational or memory cost.
We validate our approach on the task of character level language modeling where we reach state-of-the-art performances while reducing the number of FLOPS.
The code to reproduce our results is publicly available\footnote{\url{https://github.com/facebookresearch/adaptive-span}}.

\section{Approach}

\subsection{Sequential transformer network}
\label{sec:fixed}

Language modeling is the problem of assigning a probability to a sequence of tokens $(w_1,\dots,w_T)$:
$$P(w_1,\dots,w_T) = \prod_{t=1}^T P(w_t~|~w_{t-1},\dots,w_1).$$
Recent progress was made with a new auto-regressive model called Sequential Transformer~\citep{vaswani2017attention}.
A Transformer is made of a sequence of layers that are composed of a block of parallel self-attention layers followed by a feedforward network.
We refer to~\citet{vaswani2017attention} for the details on the structure.
In this paper, we make a couple of modifications to the Transformer model:
we use the relative position embeddings of \citet{shaw2018self} and the caching mechanism of~\citet{dai2018transformer} to speed up the train and test time.

\paragraph{Self-attention layer.}
A core mechanism of a transformer network is the self-attention layer, which consists of multiple attention heads working in parallel.
Each attention head applies the attention mechanism of~\citet{bahdanau2014neural} to its own input.
Given a token $t$ in a sequence, the head first computes similarities with its past, i.e., any token $r$ in the span $[t-S,t)$:
\begin{eqnarray}
  s_{tr}=\mathbf{x}_t^\top \mathbf{W}_q^\top \left( \mathbf{W}_k \mathbf{x}_r + \mathbf{p}_{t-r} \right),
\end{eqnarray}
where $\mathbf{W}_k$ and $\mathbf{W}_q$ are the ``key'' and ``query'' matrices, and $\mathbf{p}_{t-r}$ is the relative position embedding.
The attention weights are then obtained by applying a softmax function on these similarities:
\begin{eqnarray}\label{eq:att}
  a_{tr}=\frac{\exp\left( s_{tr} \right)}{\sum_{q=t-S}^{t-1}\exp\left( s_{tq}\right)},
\end{eqnarray}
Finally, the head outputs a vector $\mathbf{y}_t$ by taking the average of the past representations weighted by their attention weights:
\begin{eqnarray}
  \label{eq:output}
  \mathbf{y}_t &=& \sum_{r=t-S}^{t-1} a_{tr} \mathbf{W}_v \mathbf{x}_r,
\end{eqnarray}
where $\mathbf{W}_v$ is called the ``value'' matrix. Outputs from different heads are then concatenated together and multiplied by an output matrix $\mathbf{W}_o$ before feeding to the next layer.

Similar to the memory access mechanisms of~\citet{sukhbaatar2015end}, it pulls information from the past to update the current token representation.
Repeating this mechanism in consecutive layers allows for information to flow over long distances.
However, for each input token, each attention head scales linearly in memory and time in the context size, or attention span.
There are typically $12$ layers with $8$ heads each that processes $512$ tokens simultaneously.
This drastically limits the maximum attention span used in Transformers.

\subsection{Adaptive attention span}
\label{sec:learned}

\begin{figure}[t]
\centering
\includegraphics[width=\linewidth]{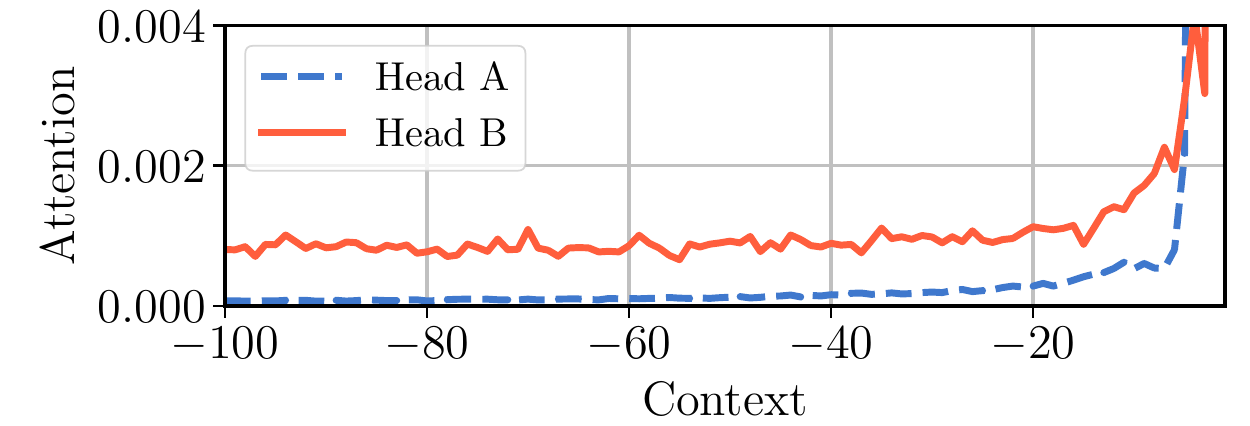}
\caption{Attention patterns of two different heads of a standard Transformer. The two patterns are qualitatively different: Head A utilizes recent steps, while Head B has uniform attention over the context.}
\label{fig:vanilla}
\end{figure}

Each attention head of a Transformer shares the same attention span $S$.
This assumes that every head requires the same span to form its representation.
As shown in Figure~\ref{fig:vanilla}, this assumption does not hold in the context of character level language modeling:
some heads (e.g., Head A) focus on the recent history, while others take information from the whole available context (e.g., Head B).
In this section, we propose to learn the attention span of each head independently to reduce their computational and memory cost.

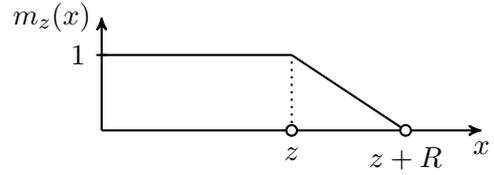
\begin{figure}[t]
\centering
  \begin{tikzpicture}[
      thick,
    >=stealth',
    dot/.style = {
      draw,
      fill = white,
      circle,
      inner sep = 0pt,
      minimum size = 4pt
    }
  ]
    \coordinate (O) at (0,0);
    \draw[->] (0,0) -- (5,0) coordinate[label = {below:$x$}] (xmax);
    \draw[->] (0,0) -- (0,1.5) coordinate[label = {left:$m_z(x)$}] (ymax);
    \draw[shift={(0,1)}] (2pt,0pt) -- (-2pt,0pt) node[left] {$1$};
    \draw[-] (0,1) -- (2.5,1) node[] {};
    \draw[-] (2.5,1) -- (4,0) node[] {};
    \draw[dotted] (2.5,0) -- (2.5,1) node[] {};
    \draw (2.5,0) node[dot, label = {below: $z$}] {};
    \draw (4,0) node[dot, label = {below: $z+R$}] {};
  \end{tikzpicture}
  \caption{The soft mask as a function of the distance.}
  \label{fig:ramp}
\end{figure}

For each head, we add a masking function to control for the span of the attention.
A masking function is a non-increasing function that maps a distance to a value in $[0,1]$.
We take the following soft masking function $m_z$ parametrized by a real value $z$ in $[0,S]$:
\begin{equation*}\label{eq:mask}
  m_z(x)=\min\left[\max\left[\frac{1}{R}\left(R+z-x\right), 0\right], 1\right],
\end{equation*}
where $R$ is a hyper-parameter that controls its softness.
This soft masking function is inspired by~\citet{jernite2016variable}.
In \fig{ramp}, we show the shape of this piecewise function as a function of the distance.
The attention weights from \eq{att} are then computed on the masked span, i.e.,
\[ a_{tr} =
  \frac{m_z(t-r)\exp\left( s_{tr} \right)}{\sum\limits_{q=t-S}^{t-1}m_z(t-q)\exp\left( s_{tq}\right)}.
\]
We add  a $\ell_1$ penalization on the parameters $z_i$ for each attention head $i$ of the model to the loss function:
\[L = -\log P(w_1,\dots,w_T) + \frac{\lambda}{M} \sum_{i} z_i,\]
where $\lambda>0$ is the regularization hyper-parameter, and $M$ is the number of heads in each layer.
Our formulation is differentiable in the parameters $z_i$ and we learn them jointly with the rest of the model.

\paragraph{Dynamic attention span.}
As an extension, we consider a dynamic computation approach~\citep{graves2016adaptive} where the attention span dynamically change based on the current input~\citep{Luong2015EffectiveAT,Shu2017AnES}.
At a time step $t$, the span parameter $z_t$ of an attention head is then a function of the input parametrized by a vector $\mathbf{v}$ and a scalar $b$, i.e., $z_t=S\sigma(\mathbf{v}^T\mathbf{x}_t+b)$.
We penalize $z_t$ in the same way as before and learn the parameters $\mathbf{v}$, $b$ jointly with the rest of the parameters.

\section{Experiments}

\begin{table*}[t]
\centering
\begin{tabular}{lcccccc}
  \toprule
  Model & \#layers & Avg. span & \#Params & \#FLOPS & dev & test \\
  \midrule
  \multicolumn{6}{l}{\emph{Small models}}\\
  T12~\citep{al2018character} & 12 & 512 & 44M & 22G & - & 1.18\\
  Adaptive-Span ($S=8192$) & 12  & 314 & 38M & 42M & 1.05 & \bf 1.11 \\
  \midrule
  \multicolumn{6}{l}{\emph{Large models}}\\
  T64~\citep{al2018character} & 64 & 512 & 235M & 120G & 1.06 & 1.13\\
  T-XL~\citep{dai2018transformer} & 24 & 3800 & 277M & 438M & - & 1.08\\
  Adaptive-Span ($S=8192$) & 24 & 245 & 209M & 179M & 1.01 & \bf 1.07 \\
  \bottomrule
\end{tabular}
  \caption{
    Character level language modeling on \texttt{text8}.
    We report bpc for the dev and test sets, as well as, the number of parameters, the average attention spans and total number of FLOPS (an estimate of the number of FLOPS necessary for computing one step prediction).
  }
\label{tab:result}
\end{table*}

In this section, we evaluate the impact of our adaptive attention mechanism in the experimental setting of~\citet{al2018character} for character level language modeling.

\paragraph{Dataset.}
We use the \texttt{text8} and \texttt{enwik8} datasets of~\citet{mahoney2011large}. The both dataset have $100$M tokens.
We report bit per character (bpc) on dev and test set.

\paragraph{Implementation details.}
We experiment with two sizes of models.
Our small models have $12$ layers and a hidden size of $d_h=512$, except for the feedforward ReLU layers, which have $2048$ units.
The large models have $24$ layers with a hidden size of $d_h=768$, and a ReLU size of $4096$. All models have $8$ attention heads in each layer.
Token and position embedding parameters are initialized from $\mathcal{N}(0,1)$, and the projection matrices $\mathbf{W}_{\{q,k,v,o\}}$ are initialized from $\mathcal{U}(-1/\sqrt{d_h}, 1/\sqrt{d_h})$.
A single set of position embeddings $\mathbf{p}_t$ is shared across all the heads.

In adaptive-span models, we reprameterized the span parameter $z$ by $z=S z' $, where $z' \in [0, 1]$ is initialized to $0$. In dynamic-span models, the bias term $b$ is initialized $-4$ to make initial spans small.
We set the hyperparameters $\lambda=2\times 10^{-6}$ and $R=32$ for the both type of models, except $\lambda$ is reduced to $0.5\times 10^{-6}$ when $S=8192$ because $z$ was not growing longer than 4000.

We use Adagrad with a batch size of $64$ and fixed learning rate of $0.07$ and $32$k warm-up steps.
Our warm-up strategy differs from~\citet{vaswani2017attention}: we linearly increase learning rate from zero to the final learning rate.
Gradients of each module are clipped at $0.03$ for better stability.
At train time, we use a block of $512$ consecutive characters and compute the loss and gradient for each of those $512$ characters.

In small models, we apply dropout with a rate of $0.3$ to the attention and the feedforward ReLU activations.
We train small models for $600K$ steps ($900K$ steps when $S=8192$), which takes about $2 \sim 3$ days on $8$ V100 GPUs depending on the attention span limit.
Large models are trained with a dropout rate of $0.4$ until the validation performance stopped improving ($250K$ steps for \texttt{text8} and $150K$ steps for \texttt{enwik8}), and then further trained for $20K$ steps with a learning rate divided by $10$.

\begin{figure*}[t]
\centering
  \begin{tabular}{lcr}
    \includegraphics[width=.31\linewidth]{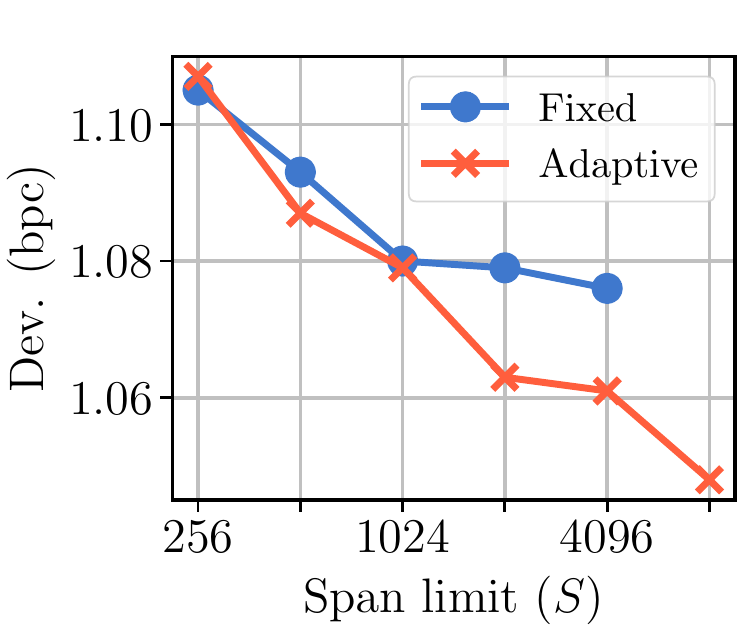}&
    \includegraphics[width=.31\linewidth]{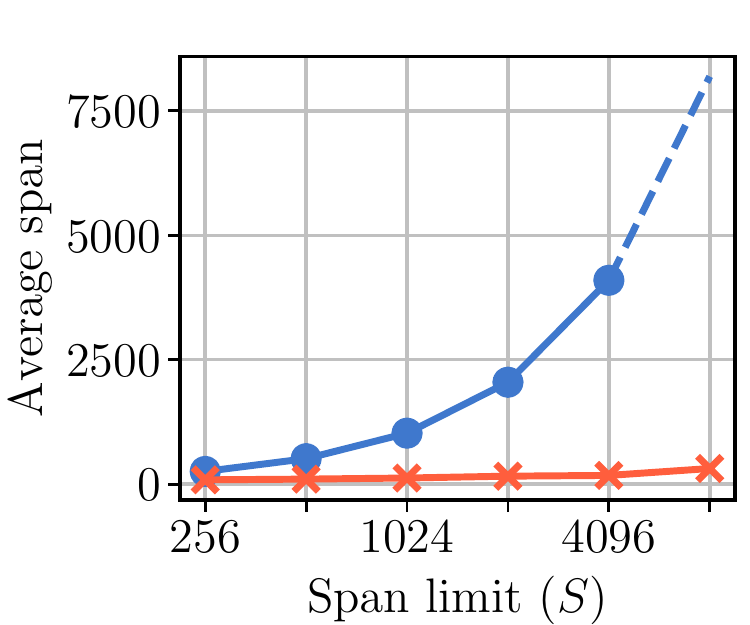}&
    \includegraphics[width=.31\linewidth]{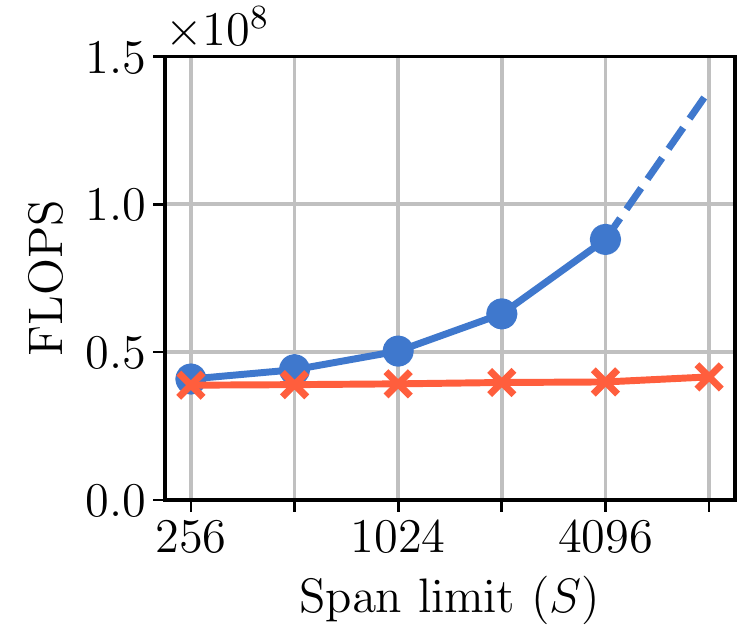}
  \end{tabular}
  \vspace{-4mm}
  \caption{\textbf{Left:} validation performances improve as the attention span limit $S$ increase (we did not train a fixed-span model with $S=8192$ due to memory limitation).
  \textbf{Center:} average attention span of trained models. Learning attention spans significantly reduces the average attention span.
  \textbf{Right:} the number of FLOPS during inference time grows almost linearly with $S$ for the fixed span models. The adaptive-span models do not have this growth in \#FLOPS because they have a very small attention span on average.}
\label{fig:perf}
\end{figure*}

\paragraph{Results.}
In \tab{result}, we compare our sequential Transformer with the adaptive spans (``Adaptive-Span'') of Sec.~\ref{sec:learned} to models of~\citet{al2018character} and \citet{dai2018transformer}.
For small models, our model outperforms the other Transformers  by $0.07$ bcp while significantly reducing the memory usage for large attention span.
Interestingly, even with a limit on span sets to $8192$, the average span is only $314$.
Similar results are obtained on \texttt{enwik8} as shown in \tab{enwiki}, where the adaptive-span model outperformed similar sized models with a significantly smaller average span. Our large models achieved state-of-the-art performances on both datasets with fewer parameters and FLOPS.

In \fig{perf}, we compare the fixed and adaptive span small Transformers as we increase the attention span limit $S$.
The performance of both models improve as the limit increase (see \fig{perf}(left)), but the adaptive-span model benefits more from longer span.
As shown on the \fig{perf}(center), a Transformer with adaptive spans controls its average spans, leading to reduction of up to $70\%$ in the number of FLOPS for the inference with large spans (see \fig{perf}(right)).

\begin{table}[t]
\centering
\setlength{\tabcolsep}{2.5pt}
\begin{tabular}{lcccc}
  \toprule
  Model & \#layers & \#Params & \#FLOPS & ~~dev / test \\
  \midrule
  \multicolumn{5}{l}{\emph{Small models}}\\
  T12 & 12 & 44M & 22G & ~~~~-~~ / 1.11 \\
  T-XL & 12 & 41M & 64M & ~~~~-~~ / 1.06 \\
  Adaptive & 12 & 39M & 41M & 1.04 / \bf 1.02 \\
  \midrule
  \multicolumn{5}{l}{\emph{Large models}}\\
  T64 & 64 & 235M & 120G & ~~~~-~~ / 1.06 \\
  T-XL & 18 & 88M & 329M & ~~~~-~~ / 1.03 \\
  T-XL & 24 & 277M & 438M & ~~~~-~~ / 0.99 \\
  Adaptive & 24 & 209M & 181M & 1.00 / \bf 0.98 \\
  \bottomrule
\end{tabular}
  \caption{
    Results on \texttt{enwik8}. The span limit is $S=8192$ for the adaptive-span models.
  }
\label{tab:enwiki}
\end{table}

\begin{figure}
\centering
\includegraphics[width=\linewidth]{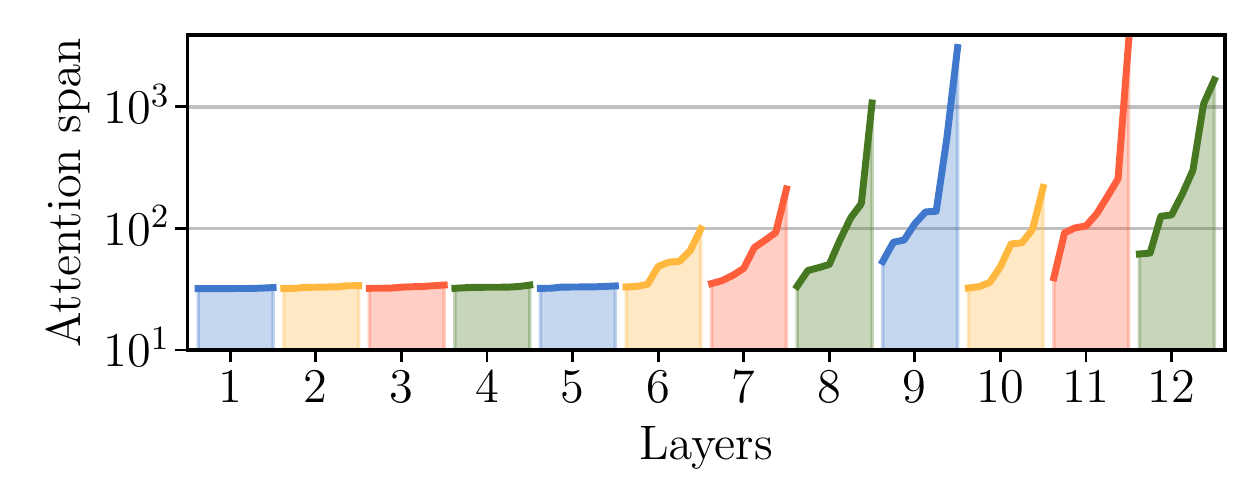} \\
\caption{
Adaptive spans (in log-scale) of every attention heads in a 12-layer model with span limit $S=4096$. Few attention heads require long attention spans.}
\vspace{-4mm}
\label{fig:span4096}
\end{figure}

\paragraph{Impact on the attention span.}
In \fig{span4096}, we show the final attention spans of every attention heads of our small adaptive-span model with $S=4096$.
Even though all the span sizes are initialized to the same value, we see large varieties in their final values.
We can see that the lowest 5 layers have the smallest possible attention span, which is $R=32$ of the masking function.
This indicates that lower layers in a Transformer model do not really require a long attention span in this particular task.
In contrast, few attention heads in the higher layers have very long spans, exceeding several thousand.
Although there is a general tendency of higher layers having longer attention spans, it is not a simple monotonic function of the layer height.

\paragraph{Impact on the number of FLOPS.}
Having a smaller attention span has a direct impact on the total number of FLOPS necessary for computing one-step prediction.
In a standard fixed-span model, the total number of FLOPS is mostly controlled by the feed-forward layer (accounting for 62\% of FLOPS when $S=256$). However, as the span increase, the attention layer dominates the computation (82\% of FLOPS when $S=8192$), making it hard to scale to longer sequences. In contrast, the learning of an attention span keeps computation at a relatively constant level even as $S$ increase as shown in \fig{perf}(right).

The memory usage is also dominated by the attention layer as the attention span increase. Thus, reducing the average span will also reduce the memory usage. However, because all heads in a single layer attend to common state vectors, the maximum span within each layer will determine the memory usage. The same is true for the number of FLOPS if all heads of a layer are computed together, as often done for better efficiency.

In practice, the largest fixed-span model that can fit in memory for training had a span of $S=2048$ (batches had to be split when $S=4096$), and it took about 550ms per batch. In contrast, an adaptive-span model with a 4 times longer span of $S=8192$ fit in memory and took about similar time per batch.

\begin{figure}[t]
\centering
\includegraphics[width=\linewidth]{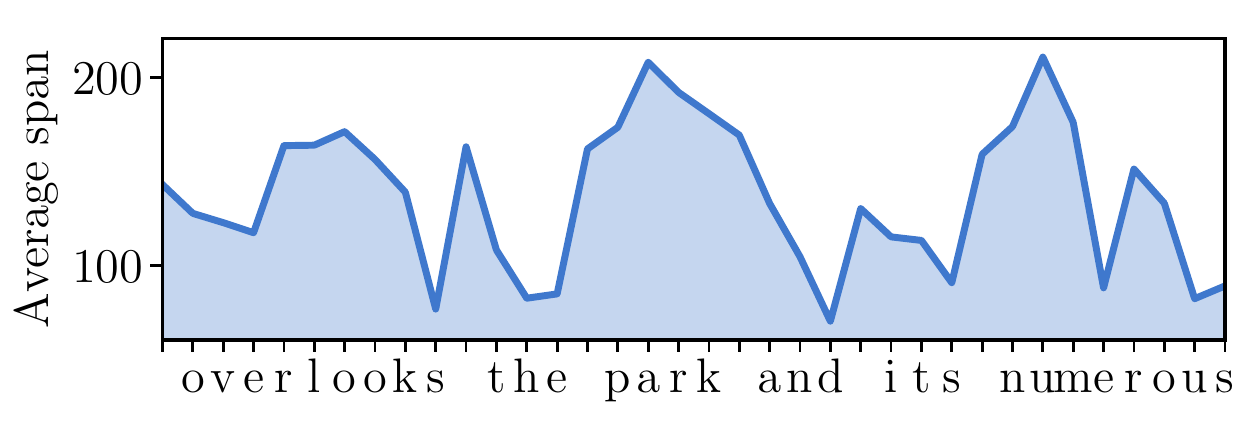}\\
\vspace{-2mm}
\caption{
  Example of average dynamic attention span as a function of the input sequence.
  The span is averaged over the layers and heads.}
\label{fig:dynamic}
\end{figure}

\begin{table}[t]
\centering
\begin{tabular}{lccccc}
  \toprule
  Model & Avg. span & dev \\
  \midrule
  Adaptive ($S=1024$) & 123 & 1.08\\
  Dynamic ($S=1024$) & 149 & 1.08 \\
  \bottomrule
\end{tabular}
  \caption{
    Comparison between adaptive and dynamic attention span on \texttt{text8}.
  }
  \vspace{-4mm}
\label{tab:dynamic}
\end{table}

\paragraph{Dynamic span.}
In Table~\ref{tab:dynamic}, we show the adaptive and dynamic spans achieved the same performance with comparable average spans on \texttt{text8}.
Figure~\ref{fig:dynamic} shows how the average dynamic span adapts to the input sequence.
The span increases at the beginning of words and in the middle of composed words, e.g., to predict the ``l'' in ``overlook''.

\section{Conclusion}
In this work, we present a novel self-attention layer with an adaptive span.
This mechanism allows for models with longer context, and thus with the capability to catch longer dependencies.
We have shown the importantce of this feature in the context of character level modeling where information is spread over great distances.


\bibliography{acl2019}
\bibliographystyle{acl_natbib}

\end{document}